\documentclass[letterpaper, 10 pt, conference]{ieeeconf}  

\IEEEoverridecommandlockouts                              
\usepackage{mathrsfs}
\usepackage{amsfonts}
\usepackage{psfig}
\usepackage{graphics} 
\usepackage{epsfig} 
\usepackage{times} 
\usepackage{amsmath} 
\usepackage{amssymb}  
\usepackage{multicol}
\usepackage{upgreek}
\usepackage{bm}
\usepackage{subfigure}
\usepackage{cite}
\usepackage{subeqnarray}
\usepackage{cases}
\usepackage{multirow}
\usepackage{times} 
\usepackage{booktabs}
\usepackage{float}
\usepackage{sansmath}
\usepackage{color}
\usepackage{comment}

\title{\LARGE \bf
Distributed Motion Control for Multiple Connected Surface Vessels}
\author{
Wei Wang, Zijian Wang$^{\ast}$, Luis Mateos$^{\ast}$, Kuan Wei Huang, Mac Schwager, Carlo Ratti, and Daniela Rus
\thanks{This work was supported by grant from the Amsterdam Institute for Advanced Metropolitan Solutions (AMS) in Netherlands.}
\thanks{W. Wang, L. Mateos, K. Huang, and C. Ratti are with the SENSEable City Laboratory, Massachusetts Institute of Technology, Cambridge, MA 02139 USA.
        {\tt\small \{wweiwang, lamateos, kwhuang, ratti\}@mit.edu}}
\thanks{W. Wang, L. Mateos, and D. Rus are with the Computer Science and Artificial Intelligence Lab (CSAIL),  Massachusetts Institute of Technology, Cambridge, MA 02139 USA.
       {\tt\small \{wweiwang, lamateos, rus\}@mit.edu}}
\thanks{Z. Wang, and M. Schwager are with the Department of Aeronautics and Astronautics, Stanford University, Stanford, CA 94305, USA.
       {\tt\small \{zjwang, schwager\}@stanford.edu}}
\thanks{$^{\ast}$These authors contributed equally to this work.}
}
\begin{document}

\maketitle
\thispagestyle{empty}
\pagestyle{empty}

\begin{abstract}
We propose a scalable cooperative control approach which coordinates a group of rigidly connected autonomous surface vessels to track desired trajectories in a planar water environment as a single floating modular structure. Our approach leverages the implicit information of the structure's motion for force and torque allocation without explicit communication among the robots.
In our system, a leader robot steers the entire group by adjusting its force and torque according to the structure's deviation from the desired trajectory, while follower robots run distributed consensus-based controllers to  match their inputs to amplify the leader's intent using only onboard sensors as feedback. To cope with the nonlinear system dynamics in the water, the leader robot employs a nonlinear model predictive controller (NMPC), where we experimentally estimated the dynamics model of the floating modular structure in order to achieve superior performance for leader-following control. Our method has a wide range of potential applications in transporting humans and goods in many of today's existing waterways. We conducted trajectory and orientation tracking experiments in hardware with three custom-built autonomous modular robotic boats, called \textit{Roboat}, which are capable of holonomic motions and onboard state estimation. Simulation results with up to 65 robots also prove the scalability of our proposed approach.

\end{abstract}
\section{Introduction}
Cooperative behavior is a remarkable phenomenon in biological systems, such as flocks of birds, schools of fish and colonies of ants. These swarm creatures can move in a coordinated fashion \cite{couzin2003self} to complete complex tasks such as foraging, transporting food and building massive structures. Some animals even physically link each other into one unit to facilitate moving in challenging environments. For example, fire ants can self-assemble into waterproof rafts for months to survive floods \cite{Mlot7669}.

Loosely inspired by these astonishing swarm behaviors, cooperative control of multi-robot systems has also been receiving increasing attention by roboticists \cite{Alonso-Mora2018} and control
theorists \cite{Olfati-Saber-733}. As it is now rapidly becoming economically viable to embed powerful computers and sensors into small robots, future vehicle could be composed of groups of physically connected smart modular robots. These modular vehicle can reconfigure and self-assemble into desired shapes and sizes to execute tasks such as transportation in certain environments. Although the complexity of multi-robot systems increases with the number of vehicle, additional vehicle can provide flexibility and robustness to handle various tasks and compensate for single vehicle failures.
This paper presents the custom hardware, dynamics model and cooperative control of a floating structure consisting of physically linked robotic modules.

Modular robots \cite{Rus5569030, Murata4141035, Yim4141032} has emerged as the forefront of robotics research since the last several decades. An intriguing property of modular robots is the ability to re-assemble, re-configure and dynamically adapt the size of the robotic fleet to the changing environment during a mission \cite{Daudelineaat4983}.
Most of the existing work in the literature focused on rigid connection among modules for aerial robots \cite{Raymond0278364913501212, Kumar8461014}, wheeled robots \cite{Baldassarre4067066}, and maritime robots \cite{paulos2015automated}. Modular structure control is similar to the problem of cooperative object transport \cite{Tuci2018a} where a group of robots are physically attached to the object and they need to be coordinated to transport objects to a final destination \cite{ Rus525802, Mellinger2013, ZijianWang7402230, ZijianWangIJRR2016}.

Some problems have yet to be fully solved on cooperative transport. First, most of the work assume a prior knowledge of the load and robot dynamics, which is not the case in real-world scenarios. Recently, a few algorithms have been proposed to estimate the inertial parameters of the load \cite{Franchi8476991, Michael7989533}. However, the damping parameters of the load are ignored in these studies.
Second, most works have focused on controlling the linear motion of the structure, while leaving the rotation uncontrolled \cite{wang2016multi, ZijianWang7487163, ZijianWang7402230}. However, rotation control is necessary in reaching an appropriate orientation to navigate through spatially constrained environments. Third, current work within the leader-follower scope typically focuses on the design of the follower controller, while the leader controller is always simple and doesn't consider the robot dynamics. These model-free leader controllers cannot adapt to different sizes and configurations of the group and will not work well in highly nonlinear circumstances such as the water environment. The leader controller is critical in stabilizing and optimizing the task performance of the group. Fourth,  current cooperative algorithms have rarely considered dynamic systems on the water \cite{Esposito4543414, Hajieghrary8430951, Hajieghrary8206033}.  Cooperative transport using a team of surface vehicles poses unique challenges not encountered in aerial or ground vehicles. For instance, inertia of the vehicles and the load become more significant factors leading the system harder to control.

Based on the above discussion, we propose a novel solution for cooperative control of a floating modular structure in this paper. We require only one leader robot in the group to know the desired trajectory and orientation, and the leader can steer the rest of the robots by adjusting its input. It is noted that our approach requires no communication between any two robots and yet all the robots will contribute positively to both the structure's translation and rotation motion. Moreover, robots ignore relative positions in the structure and treat their local position and velocity as if being measured at the center of the structure. This reduces the amount of information that the designer has to provide to the robots about the modular structure and avoids providing robots with information about their positions in the structure.

More specifically, the contributions of this paper are outlined as follows.
First, a new NMPC strategy is proposed for the leader which can adapt to different group sizes and configurations without re-tuning the parameters.
Second, we derive a dynamical model for the rigidly attached modular structure with $N$ robots in different configurations, which serves as a realistic input rather than an assumption for the cooperative algorithm.
Third, we extend prior works \cite{wang2016multi, ZijianWang7487163, ZijianWang7402230} which only considered translation control by incorporating rotation sensing and torque input, so that we can control the translation and orientation of the modular structure independently. Note that rotation control is also designed in \cite{wang2018ouijabots}. However, the algorithm requires every robot to know its own position relative to the center of mass (CoM) of the structure, while our algorithm ignores the robot's relative position to the CoM, and only needs local measurements for each robot. Fourth, we develop three autonomous robotic boats which are capable of onboard state estimation. The cooperative algorithms are verified by both the simulations and experiments with these surface modules. Some of our early results were briefly summarised as an extended two-page abstract in a conference workshop \cite{wang2019cooperative}, and this paper adds substantial new contributions including the theoretical results, the NMPC formulation, controller analysis and extensive simulation studies.

\section{Modular Robot Design} \label{ModuleDescription}
Our design focuses on developing a modular robotic boat with the ability to estimate its pose, velocity and acceleration using onboard sensors.  Its main components are described as follows.
\subsection{Surface Vessel}
\begin{figure}[htb]
    \centering
    \includegraphics[width=1.0\linewidth] {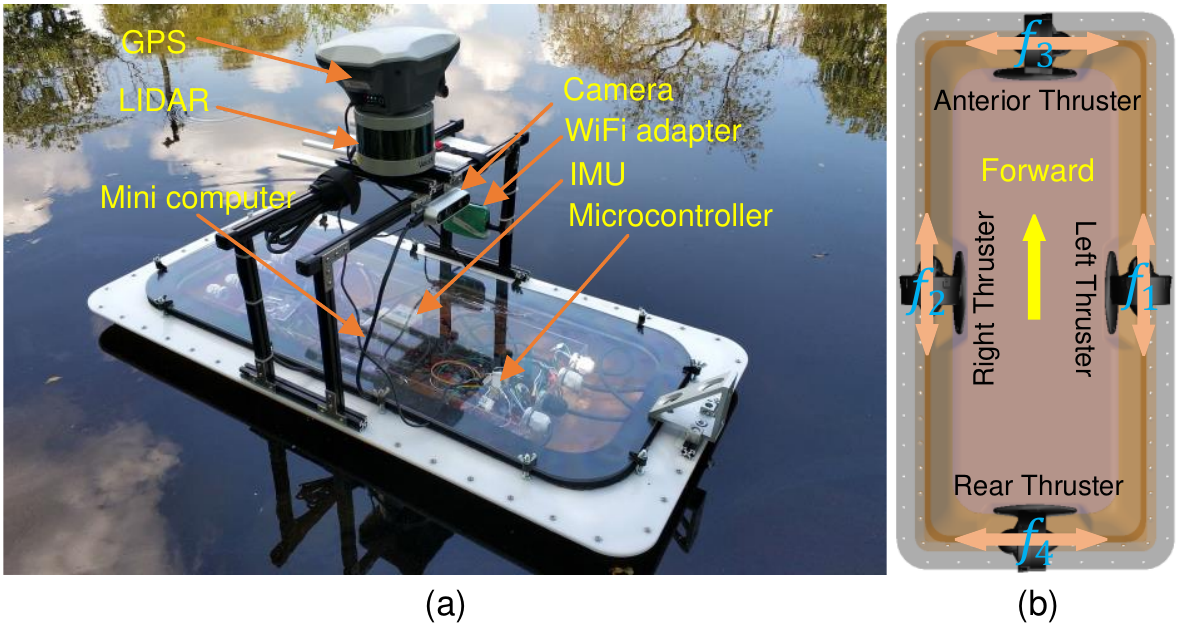}
    \caption{The  modular robotic boat. (a) The prototype picture (side view); (b) thruster configuration (bottom view). $f_1$, $f_2$, $f_3$ and $f_4$ denote the forces generated by left, right, anterior and rear thrusters, respectively.}
    \label{RoboatDesign3}
\end{figure}

As shown in Fig. \ref{RoboatDesign3}, the robot has four thrusters around the hull to achieve holonomic motions.
In our latest hardware iteration, we mold the boat hull with fiber glass to improve its strength and water resistance.
An Intel NUC is adopted as the main controller running the Robotic Operating System (ROS). Moreover, an auxiliary microprocessor (STM32F103) is used for real-time actuator control. The robot has multiple onboard sensors including a 3D LiDAR (Velodyne, Puck VLP-16), an IMU, a camera and a GPS sensor. The camera and GPS are not used in this paper.
The vessel weighs around 15 kg, and its dimensions are 0.90 $\times$  0.45 $\times$ 0.15 m.  It is powered by a 11.1 V Li-Po battery which lasts around three hours.  Each  propeller  is  fixed  and  can  generate  forward  and backward forces. More details of the robot hardware can be found in \cite{WeiICRA2018}.
\subsection{Onboard State Estimation}
We use an extended Kalman filter (EKF) \cite{bishop2001introduction} to fuse LiDAR with IMU measurements and obtain stable and precise state estimation for our surface vessels.
In particular, we select LiDAR as the primary sensor for estimation because of its accuracy. Moreover, we choose the normal distributions transformation (NDT) matching algorithm \cite{BiberStrasser2006} because it handles noisy measurements well thanks to its probabilistic modeling.
The EKF algorithm \cite{moore2016generalized} further integrates the outputs of the NDT algorithm and IMU odometry to estimate the state (including position, velocity and acceleration) of the surface vessel.  More details of the EKF design for our vessel can be referred as to our previous studies \cite{WeiIROS2019}.


\section{System Modeling} \label{ModuleDynamics}
The dynamics equations of the robotic boat while in motion in the water environment are given in this section. We also connect multiple robots to form a rigid structure to achieve coordinated trajectory tracking without explicit communication among the robots.

\noindent \textbf{Definition 1. (Modular Vessel)}: A robotic boat that can 1) move on the water surface by itself and 2) estimate its states including pose, velocity and acceleration by onboard sensors.

\noindent \textbf{Definition 2. (Floating Modular Structure)}: A group of $N$ rigidly connected modular robots that move on the water as a single rigid body in two dimensional space. All modules are homogeneous, including shape, mass, inertia, sensors and actuators.

\begin{figure}[htb]
    \centering
    \includegraphics[width=0.85\linewidth] {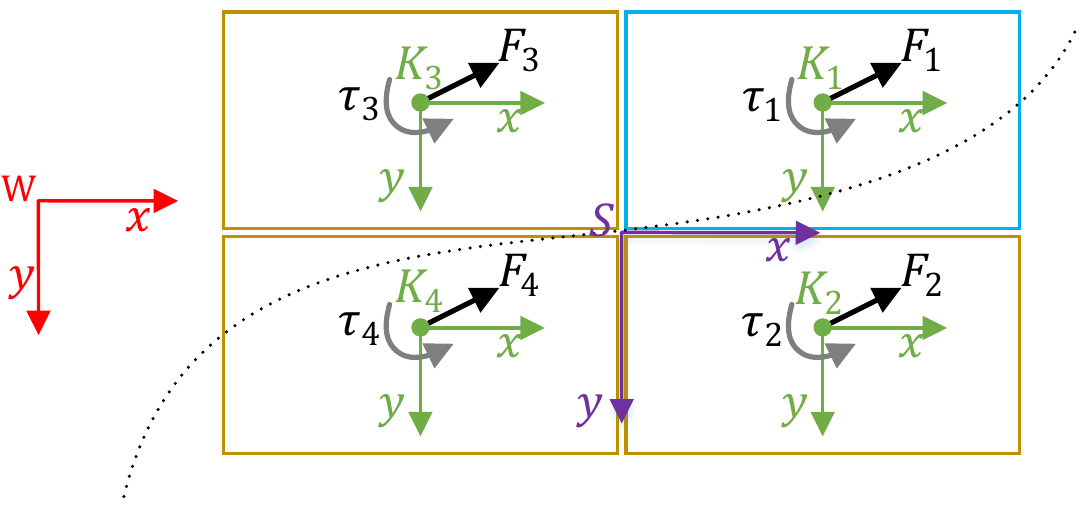}
    \caption{Representation of a structure with four modules. The black and gray arrows represent the force and torque respectively. The red, purple, and green axes represent the world, structure and module frames respectively.}
    \label{MultiRobotFrame}
\end{figure}

Three coordinate systems are used to describe the modular vessel and the rigidly connected floating modular structure, as illustrated in Fig. \ref{MultiRobotFrame}.
$W$ denotes the inertial coordinate frame which has its $z$-axis pointing downwards. The body-fixed $K_i$ denotes the module coordinate frame,  whose origin is the center of mass of the vessel.  $S$ denotes the structure coordinate frame, whose origin is attached to the center of mass of the floating structure. Without loss of generality, we assume that the headings of all modules are aligned when connected in the structure as depicted in Fig. \ref{MultiRobotFrame}.
\subsection{Dynamical Model of a Modular Vessel}

The kinematic equation correlates the velocity components in the inertial frame to those in the body-fixed frame, and the equation is described as follows
\begin{eqnarray}\label{kinematicequation}
\dot{\bm{\eta}}=\bm{T}(\bm{\eta})\bm{v}
\end{eqnarray}
where $\bm{\eta}=[x~y~\psi]^{T}\in\mathbb{R}^{3\times1}$ denotes the position and orientation of the vessel in the inertial frame and $\bm{T}(\bm{\eta})\in\mathbb{R}^{3\times3}$ is the transformation matrix converting a state vector from the body-fixed frame to the inertial frame; $\bm{v}=[u~v~\omega]^{T}\in\mathbb{R}^{3\times1}$ denotes the vessel velocity, which contains the vessel surge velocity ($u$), sway velocity ($v$) and angular velocity ($\omega$) in the body-fixed frame;

The dynamics of a surface vessel can be described by a nonlinear differential equation \cite{Fossen285}
\begin{eqnarray}\label{GeneralUSVDynamics}
\bm{M}\dot{\bm{v}}+\bm{C}(\bm{v})\bm{v}+\bm{D}(\bm{v})\bm{v}=\bm{d}
\end{eqnarray}
where $\bm{M }\in\mathbb{R}^{3\times3}$ denotes the positive-definite symmetric added mass and inertia matrix; $\bm{C}(\bm{v})\in\mathbb{R}^{3\times3}$ denotes the skew-symmetric vehicle matrix of Coriolis and centripetal terms; $\bm{D}(\bm{v})\in\mathbb{R}^{3\times3}$ denotes the positive-semi-definite drag matrix-valued function, $\bm{d} \in\mathbb{R}^{3\times1}$  contains the 2D forces $\bm{F}$  and the 1D torque $\tau$ applied to the vessel.
$\bm{M}$ is defined as: $\bm{M}=\text{diag}\{m_{11}, m_{22}, m_{33}\}$, and $\bm{C}(\bm{v})$ is expressed as
\begin{eqnarray}\label{CoriolisMaxtrix}
\bm{C}(\bm{v})=
 \left[
 \begin{array}{ccc}
0                                   &  0                               & -m_{22}v\\
0                                   &0                                 & m_{11}u\\
m_{22}v               &-m_{11}u               &0
\end{array}
\right]
\end{eqnarray}
Furthermore, the drag matrix $\bm{D}(\bm{v})$ is simplified as $\bm{D}(\bm{v})=\text{diag} \{X_u, Y_v, N_{\omega}\}$ due to the low speed of the vessel in experiments.
Furthermore, the applied force and torque vector $\bm{d}$ can be written as
\begin{eqnarray}\label{AppliedForceMaxtrix}
\bm{d}
=\bm{B}\bm{u}
=
\left[
 \begin{array}{cccc}
1                      &  1                                      &    0                       & 0\\
0                      &  0                                      &    1                       & 1\\
\dfrac{a}{2}&-\dfrac{a}{2}                 &     \dfrac{b}{2} &-\dfrac{b}{2}
\end{array}
\right]
 \left(
 \begin{array}{c}
f_1\\
f_2\\
f_3\\
f_4
\end{array}
\right)
\end{eqnarray}
where $\bm{B}$ is the control matrix describing the thruster configuration and $\bm{u}$ is the
control input. $a$ is the distance between the transverse propellers, $b$ is the distance between the longitudinal propellers. $a$ and $b$ are also the length and width of the vessel, respectively. As shown in Fig. \ref{RoboatDesign3}(b), $f_1$, $f_2$, $f_3$ and $f_4$ represent the forces generated by the left, right, anterior and rear propellers, respectively.
Finally, (\ref{GeneralUSVDynamics}) is rewritten as
\begin{eqnarray}\label{MFCPredynamics}
&&\dot{\bm{v}}=\bm{M}^{-1}\bm{B}\bm{u}-\bm{M}^{-1}(\bm{C}(\bm{v})+\bm{D}(\bm{v}))\bm{v} \label{MFCPredynamicsB}
\end{eqnarray}
The unknown hydrodynamic parameters $\bm{M}$, $\bm{C}$, and $\bm{D}$ are identified by a nonlinear least squares method based on the trust-region-reflective algorithm \cite{WeiICRA2018}.

\subsection{Modeling of Floating Modular Structure}\label{DynamicsOfFloatingModularStructure}
The floating modular structure is composed of a group of rigidly attached $N$ robots, which are labelled as \{$R_1, R_2, ... , R_N$\}. We assign $R_1$ as the leader robot and the rest as the follower robots. We assume that only the leader $R_1$ knows the desired trajectory (e.g., position, orientation, etc.), while the followers have no information about the desired trajectory. Moreover, none of the robots within the group and are permitted to communicate with each other. Furthermore, no robot knows its relative positions in the group.

\begin{figure}[htb]
    \centering
    \includegraphics[width=0.9\linewidth] {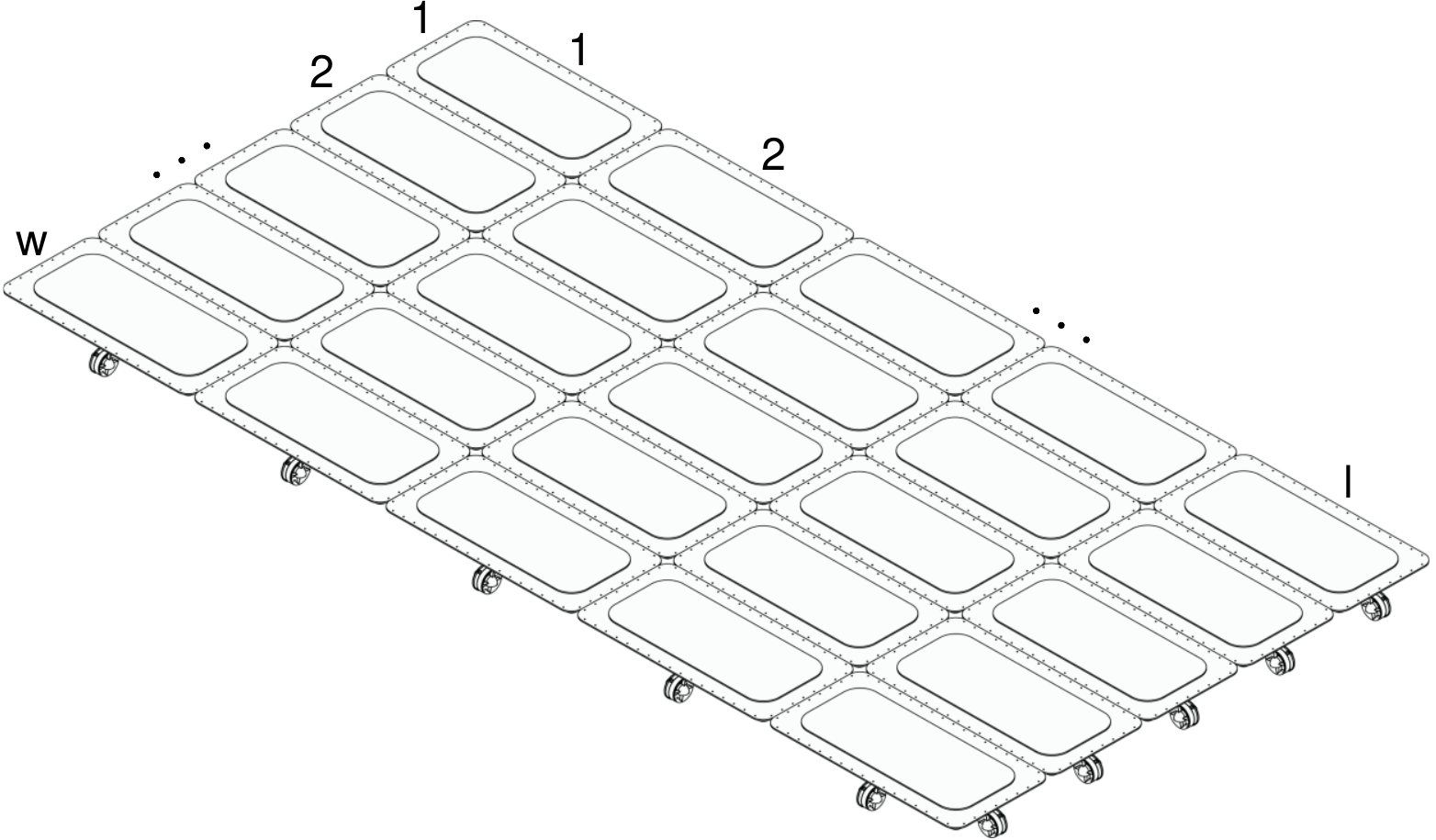}
    \caption{Diagram of the modular structure which has $l$ robots in longitudinal direction and $w$ robots in lateral direction. }
    \label{roboatgroupstructure}
\end{figure}

We now derive a dynamical model for the floating modular structure which contains $N$ rigidly connected vessels. The dynamical model of the floating structure is also described by a nonlinear differential equation
\begin{eqnarray}\label{NRobotDynamics}
\bm{M}_g\dot{\bm{v}}+\bm{C}_g(\bm{v})\bm{v}+\bm{D}_g(\bm{v})\bm{v}=\bm{d}_g
\end{eqnarray}
where $\bm{M}_g\in\mathbb{R}^{3\times3}$, $\bm{C}_g\in\mathbb{R}^{3\times3}$ and $\bm{D}_g\in\mathbb{R}^{3\times3}$ represent the added mass and inertia matrix, the Coriolis and centripetal matrix and the drag matrix, respectively for the modular structure; $\bm{d}_g\in\mathbb{R}^{3\times1}$ represents the total force and torque applied to the structure.
For simplicity, we assume the shape of the modular structure as solid cuboid in this study. Moreover, we define the number of the robot $N$ as: $N=l\cdot w$ where $l$ is the number of the robot along the longitudinal direction of the structure, and $w$ is the number of the robot along the lateral direction of the structure, as shown in Fig. \ref{roboatgroupstructure}.
That is, the length of the floating platform is $l\cdot a$, and the width of the floating platform is $w\cdot b$.

In particular, $\bm{M}_g$ is roughly derived from the definition of mass and moment of inertia, and it is described as follows
\begin{eqnarray}\label{Nmassmatrix}
\bm{M}_g
=\text{diag} \{Nm_{11}, Nm_{22}, \dfrac{m_{33}(l^2a^2+w^2b^2)}{12}\}
\end{eqnarray}
Note that in (\ref{Nmassmatrix}) we linearly scale the inertia tensor for the modular structure for simplicity. More accurate estimation should be investigated in the future.
Then, according to  (\ref{CoriolisMaxtrix}) and (\ref{Nmassmatrix}),  $\bm{C}_g(\bm{v})$ is defined mathematically as follow
\begin{eqnarray}\label{NCoriolis}
\bm{C}_g(\bm{v})=N \bm{C}(\bm{v}),
 \end{eqnarray}
Further, according to the definition of linear and angular drag in a perfect fluid \cite{nakayama2018introduction},  the damping of the floating structure is derived as follow
\begin{eqnarray}\label{Ndrag}
\bm{D}_g(\bm{v})
=\text{diag} \{lX_u, wY_v, \dfrac{l^3N_{\omega}}{4}\}
\end{eqnarray}
Finally, the total force and torque $\bm{d}_g$ is described as follow
 \begin{eqnarray}\label{NForce}
\bm{d}_g=
\sum_{i=1}^{N}\bm{d}_i+ [0~0~ ||\sum_{i=1}^{N}\bm{r}_i \times \bm{F}_i||]^{T},
\end{eqnarray}
where $\bm{d}_i=[ \bm{F}_i~ \tau_i]^{T}\in\mathbb{R}^{3\times1}$ denotes the 2D force and direct 1D torque applied by $R_i$, and $\bm{r}_i $ denotes the position vector pointing from the center of the mass of the structure to the center of mass of $R_i$.

\section{Control of Floating Modular Structure}
In this section we present our decentralized leader-following controller which guarantees the coordinated motion among all robot modules without explicit communication. A few assumptions are required for the robots to achieve the force and torque coordination with no communication. They are stated formally below.

\noindent \textbf{Assumption 1.}  \textit{All robots know the number of the vehicles ($N$), the length of the floating structure ($la$) and the width of the structure ($wb$)};

\noindent \textbf{Assumption 2.}  \textit{The center of mass of the robots are centrosymmetric around the center of mass of the floating structure, meaning that for any robot $R_i$, there exists another robot $j \neq  i$ such that $\bm{r}_i = -\bm{r}_j$.}

\subsection{Decentralized Follower's Controller} \label{DistributedFollowerController}

The following force and torque controller is used by every follower robot which will lead to synchronization to the leader's input:
\begin{eqnarray}\label{DistributedControlLaw}
\dot{\bm{d}}_{i} &=&\sum_{j=1, j\neq i}^{N} (\bm{d}_{j}-\bm{d}_{i})=\sum_{j=1}^{N} \bm{d}_{j}-N\bm{d}_{i} \label{ConsensusLaw} \\
&=&\bm{M}_g\dot{\bm{v}}+\bm{C}_g(\bm{v})\bm{v}+\bm{D}_g(\bm{v})\bm{v}-N\bm{d}_i. \label{ConsensusLawWithoutCommunication}
\end{eqnarray}
(\ref{ConsensusLaw}) is a notable implementation of the famous consensus law \cite{Olfati-Saber-733}
and will lead to all robots applying equal forces and torques to the modular structure. Traditional use of this protocol would require robot $i$ to communicate with its neighbors to obtain $\bm{d}_{j}$. Our novel implementation of the consensus protocol in (\ref{ConsensusLawWithoutCommunication}) eliminates the need for communication, because the sum of all neighbors' $\bm{d}_{j}$  plus robot $i$'s own $\bm{d}_{i}$ is a known quantity by (\ref{NRobotDynamics}), assuming that robot $i$ can measure $\bm{v}$ and $\dot{\bm{v}}$. Therefore the follower's control law (\ref{ConsensusLawWithoutCommunication}) can be calculated locally without communication. Note that we do not consider the second term $[0~0~||\sum_{j=1}^{N}\bm{r}_j \times \bm{F}_j||]$  of (\ref{NForce}) in the distributed controller (\ref{DistributedControlLaw}) because this term will diminish to zero when (\ref{DistributedControlLaw}) converges under Assumption 2.

\noindent \textbf{Convergence Analysis for (\ref{ConsensusLawWithoutCommunication}).} Olfati-Saber and Murray in \cite{Olfati-Saber-733} proved that the values of the all followers will converge to the leader's if there is one leader who does not change its value. If the leader's input force is changing, we can prove that the the followers' forces converge to the leader's exponentially using the similar method described by (31) in \cite{ZijianWangIJRR2016}. The convergence for a varying leader's torque will be analyzed by (\ref{RotationControlWithLocalMeasurement}) in the following. $\hfill{\Box}$

In practice, it is unrealistic for the followers to directly measure $\dot{\bm{v}}$ and $\bm{v}$ in (\ref{ConsensusLawWithoutCommunication}) at the center of mass of the structure. Here we show that the control law still works when the followers use the local measurements instead of the unrealistic $\dot{\bm{v}}$ and $\bm{v}$. The control law with local sensing is described as
\begin{eqnarray}\label{DistributedControlLawWithLocalMeasurement}
\dot{\bm{d}}_{i} &=&\bm{M}_g\dot{\bm{v}}_i+\bm{C}_g(\bm{v}_i)\bm{v}_i+\bm{D}_g(\bm{v}_i)\bm{v}_i-N\bm{d}_i,
\end{eqnarray}
where $\bm{v}_i=[u_i~v_i~\omega_i]^{T}$ is the local velocity of $R_i$. After $\bm{d}_i$ is obtained, the propeller input $\bm{u}_i$ of $R_i$ can be calculated using (\ref{AppliedForceMaxtrix}). Note that (\label{DistributedControlLawWithLocalMeasurement}) ignores the relative positions of the robots in the group.

\noindent \textbf{Convergence Analysis for (\ref{DistributedControlLawWithLocalMeasurement}).} We first analyze the convergence for the rotation control. Extracting the third row in (\ref{DistributedControlLawWithLocalMeasurement}), the control law for rotation can be written as
\begin{eqnarray}\label{RotationControlWithLocalMeasurement}
\! \! \!\! \! \! \! \! \! \!\dot{\tau}_{i} =  \! \! \!\! \! \! \! \! \! \!&&M_g^{33}\dot{\omega}+Nm_{11}(u_c+\omega  r_i^{y})- \nonumber\\
\! \! \!\! \! \! \! \! \! \! && Nm_{22}(v_c+ \omega  r_{i}^{x})+D_g^{33} r-N\tau_i \nonumber \\
\! \! \!\! \! \! \! \! \! \!=  \! \! \!\! \! \! \! \! \! \! && M_g^{33}\dot{\omega}+Nm_{11}u_c-Nm_{22}v_c +D_g^{33} r-N\tau_i+ \nonumber\\
 \! \! \!\! \! \! \! \! \! \!&& N(m_{11} \omega  r_i^{y}-m_{22} \omega  r_{i}^{x}) \nonumber \\
\! \! \!\! \! \! \! \! \! \! =  \! \! \!\! \! \! \! \! \! \! && (\sum_{j=1}^{N} \tau_{j}-N\tau_{i}) + N(m_{11} \omega  r_i^{y}-m_{22} \omega  r_{i}^{x})
\end{eqnarray}
where $M_g^{33}$  is the element in the third row and third column of $\bm{M}_g$, $D_g^{33}$ represents the same element of $\bm{D}_g$, $u_c$ and $v_c$ are the surge and sway speeds at the center of the structure respectively, $r_i^{x}$ and $r_i^{y}$ are the the $x$ and $y$ component of $\bm{r}_i$ respectively.
The first term in (\ref{RotationControlWithLocalMeasurement}) by itself would lead to a consensus, as in (\ref{ConsensusLaw}). The second term in (\ref{RotationControlWithLocalMeasurement}) can be regarded as disturbance, which denotes the additional Coriolis and centripetal effects caused by the rotation. However, under Assumption 2, the second term in (\ref{RotationControlWithLocalMeasurement}) will not have any influence on the consensus because it will be canceled out by the paired symmetric robots.

Now we analyze the convergence  of the translation control in $x$ axis. Extracting the first row in (\ref{DistributedControlLawWithLocalMeasurement}), the control law can be written as follow
\begin{eqnarray}\label{XTranslationControlWithLocalMeasurement}
\dot{F}_{i}^{x} =  \! \! \!\! \! \! \! \! \! \!&&M_g^{11}(\dot{u}_c+ \dot{\omega} r_i^{y}+\omega  (\omega r_i^y))+ \nonumber\\
 &&Nm_{22}(u_c+\omega  r_i^{y})(v_c+\omega  r_i^{x})+ \nonumber\\
 && D_g^{11}(u_c+\omega  r_i^{y})-NF_i^{x} \nonumber \\
=  \! \! \!\! \! \! \! \! \! \! && (M_g^{11}\dot{u}_c+Nm_{22}u_cv_c+D_g^{11}u_c-NF_i^x)+\dot{\omega} r_i^{y}+\nonumber\\
 &&Nm_{22}(u_c\omega r_i^x+v_c\omega r_i^y+(\omega r_i^y)( \omega r_i^x) )+\nonumber\\
&& D_g^{11} \omega  r_{i}^{y}+\omega  (\omega r_i^y)
\end{eqnarray}
Similar to (\ref{RotationControlWithLocalMeasurement}), the first term in (\ref{XTranslationControlWithLocalMeasurement}) equals to $\sum_{j=1}^{N} F_{j}^x-NF_{i}^x$ which would lead to a consensus, and the other terms in (\ref{XTranslationControlWithLocalMeasurement}) are treated as disturbances. Under Assumption 2, the third, fourth and fifth terms in (\ref{XTranslationControlWithLocalMeasurement}) will be canceled out by the paired symmetric robots.
 We need to further study the second term in (\ref{XTranslationControlWithLocalMeasurement}). Using (\ref{NRobotDynamics}), we can write $\dot{\omega}$  as follow
\begin{eqnarray}\label{omegadynamics}
\dot{\omega}= &&\dfrac{\tau_g}{M_g^{33}}+\dfrac{||\sum\limits_{j=1}^{N}\bm{r}_j \times \bm{F}_j||}{M_g^{33}} + \\ \nonumber
&&\dfrac{\omega(Nm_{22}v_c-Nm_{11}u_c-D_g^{33})}{M_g^{33}}
\end{eqnarray}
The first and third term in (\ref{omegadynamics}) will be cancelled out under Assumption 2 after the cross product with $r_i^y$. The second term $||\sum_{j=1}^{N}\bm{r}_j \times \bm{F}_j||$ will diminish to zero when the force convergence is achieved under Assumption 2. Note that Assumption 2 does not guarantee that this term will always be zero especially during the turning process. In this case, we treat the torque term as the disturbance, which can be effectively rejected by our controller.
The convergence analysis for the translation control in the $y$ axis is the same as that of the $x$ axis. $\hfill{\Box}$
\subsection{Leader's Controller}
The leader controller largely affects the tracking performance of the group. An optimal leader controller is desired to minimize the tracking errors while conforming to the constraints of the dynamics and actuator limitations.
We revised a NMPC strategy \cite{WeiICRA2018} for the leader which could adapt to various sizes and configurations without re-tuning the parameters.

The NMPC is used to track the whole state of the leader $\bm{q}_1$ where $\bm{q}_1=[x_1~y_1~\psi_1~u_1~v_1~\omega_1]^{T}$. The reference state for the leader is defined as $\bm{q}_r=[x_r~y_r~\psi_r~u_r~v_r~\omega_r]^{T}$. Then, the tracking error between the actual and reference states is defined as $\bm{e} = \bm{q}_1-\bm{q}_r$.
By combining (\ref{kinematicequation}) and (\ref{MFCPredynamicsB}), the dynamical model of the leader is then reformulated as
\begin{eqnarray}\label{MPCdynamics}
\dot{\bm{q}}_1=g(\bm{q}_1,\bm{u}_1)
\end{eqnarray}

The NMPC leader generates feasible inputs to guide the group to follow a set of predefined states. In particular, the input applied to the system is given by the solution to the following receding horizon optimal control problem, which is solved at every time step,
\begin{eqnarray}\label{ObjectiveFunction}
\underset{\bm{q}_1(\cdot), \bm{u}_1(\cdot)}{\text{min}}=\int_t^{t+T}F(\bm{q}_1(\tau),\bm{u}_1(\tau))d\tau+E(\bm{q}_1(t+T))
\end{eqnarray}
subject to
\begin{eqnarray}\label{NMPCSubject1}
\dot{\bm{q}}_1(\tau)=g(\bm{q}_1(\tau),\bm{u}_1(\tau)),~\bm{q}_1(t)=\bm{q}_1^{t},
\end{eqnarray}
\begin{eqnarray}\label{NMPCSubject2}
\bm{q}_1(\tau)\in Q_1, \bm{u}_1(\tau)\in U_1, \forall \tau \in [t, t+T]
\end{eqnarray}
 where $T$ is the prediction horizon, $Q_1$ and $U_1$ are the state space and actuator force space respectively, $F$ is the cost function defining the desired performance objective and $E$ is the terminal cost.
In particular, $F$ and $E$ are defined as follows
\begin{subequations} \label{StageCostFunction}
\begin{eqnarray}
F(\bm{q}_1,\bm{u}_1)&=&\bm{e}(t)^TQ\bm{e}(t)+\bm{u}_1(t)^{T}R\bm{u}_1(t)\\
E(\bm{q}_1)&=&\bm{e}(t+T)^TQ_{N}\bm{e}(t+T)
\end{eqnarray}
\end{subequations}
where $Q$ and $R$ are the positive definite weight matrices that penalize deviations from the desired values; $Q_N$ is the terminal penalty matrix which can improve the stability of the NMPC algorithm.
To adapt the leader controller to different sizes and configurations,we experimentally found a rough relationship between the fleet size and the good weight parameters $Q$ , $R$ and $Q_N$: $Q=NQ_m$, $R=R_m$ and  $Q_N=NQ_{N_m}$ where $Q_m$, $R_m$ and $Q_{N_m}$ are the weight parameters for an individual robot.

\section{Simulations}
Simulations are conducted in Matlab to verify the proposed distributed controller for the followers and the NMPC for the leader. ACADO Model Predictive Control Toolkit \cite{houska2012acado} is used to solve the constrained nonlinear optimization problem in the NMPC. The numerical Runge-Kutta-Fehlberg  (RKF45) method is adopted to calculate the distributed force and torque controller. The numerical Gauss-Legendre integration method is employed to simulate the dynamics of the floating modular structure.  The sampling rate of the simulation is set to 10 Hz.

In the simulation, the modular structure consists of 65 robots. The 64 followers are arranged into a 8$\times$8 array and the leader is attached to front of the array.
The dimensions and the dynamics of the simulated robots are the same as the physical roboat described in Section \ref{ModuleDescription}. According to the propeller configuration and force limitation for each propeller, each robot can apply a force up to 16.97 N and a torque up to 8.10 N$\cdot$m.
Noisy sensing and actuation are added to the simulated robots. The measured velocities of the structure at the local frame of the robots, are corrupted by zero-mean Gaussian noise $\mathcal{N}( 0, 0.1I_3)$, where $I_3$ is a three-dimensional identity matrix. Similarly, we add $\mathcal{N}( 0, 0.1I_3)$  to the desired force and torque actuation, and add $\mathcal{N}( 0, 0.03I_3)$  to the local acceleration measurements. The goal is to navigate the floating structure through a narrow canal where both the position and orientation of the structure are controlled at all times to avoid collision with the canal edges. Moreover, the desired velocities of the structure are also controlled during the whole process. The NMPC leader controller in (\ref{ObjectiveFunction})-(\ref{NMPCSubject2}) guides the whole structure to track these desired states of the structure. The parameters and constraints of the leader controller are listed as follows: $f^{\text{min}}_{i} \leq f_i \leq f^{\text{max}}_{i}$, where $f^{\text{min}}_{i} = -6$ N, $f^{\text{max}}_{i}=6$ N, and $i=1,~2,~3,~4$; $H = 4$ s, $N=65$ $Q_m=\text{diag}\{20, 20, 20, 40, 40, 40\}$, $R_m=\text{diag}\{1,1,1,1\}$, and $Q_{Nm}=\text{diag}\{20, 20, 20, 40, 40, 40\}$.

\begin{figure}[htb]
    \centering
    \includegraphics[width=1.0\linewidth] {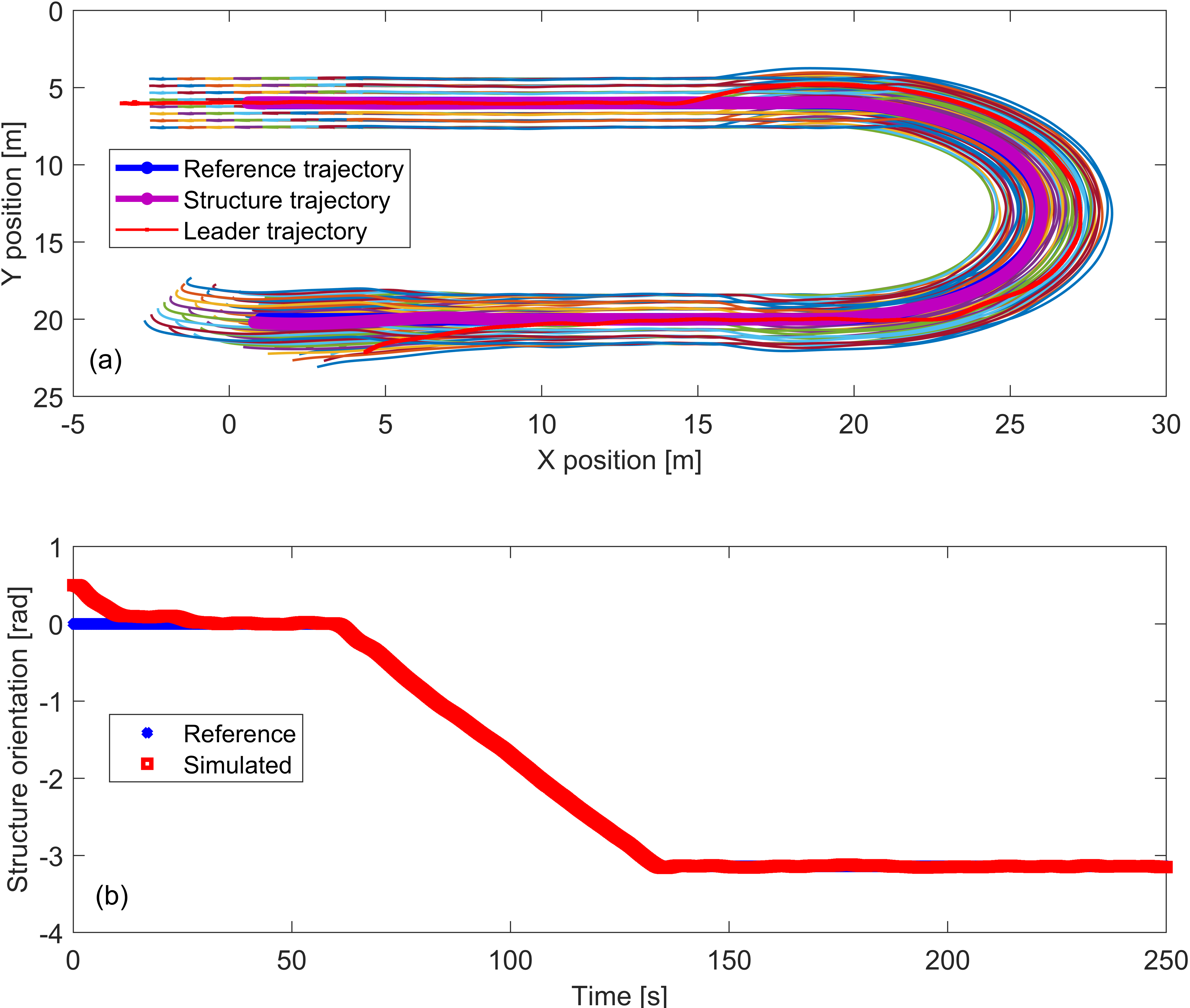}
    \caption{The trajectory and orientation of the structure in the simulation. (a) The trajectory, where the thin colored lines are the trajectories of the followers; (b) the orientation. }
    \label{SimulationTrajectoryandOrientation}
\end{figure}

\begin{figure}[htb]
    \centering
    \includegraphics[width=1.0\linewidth] {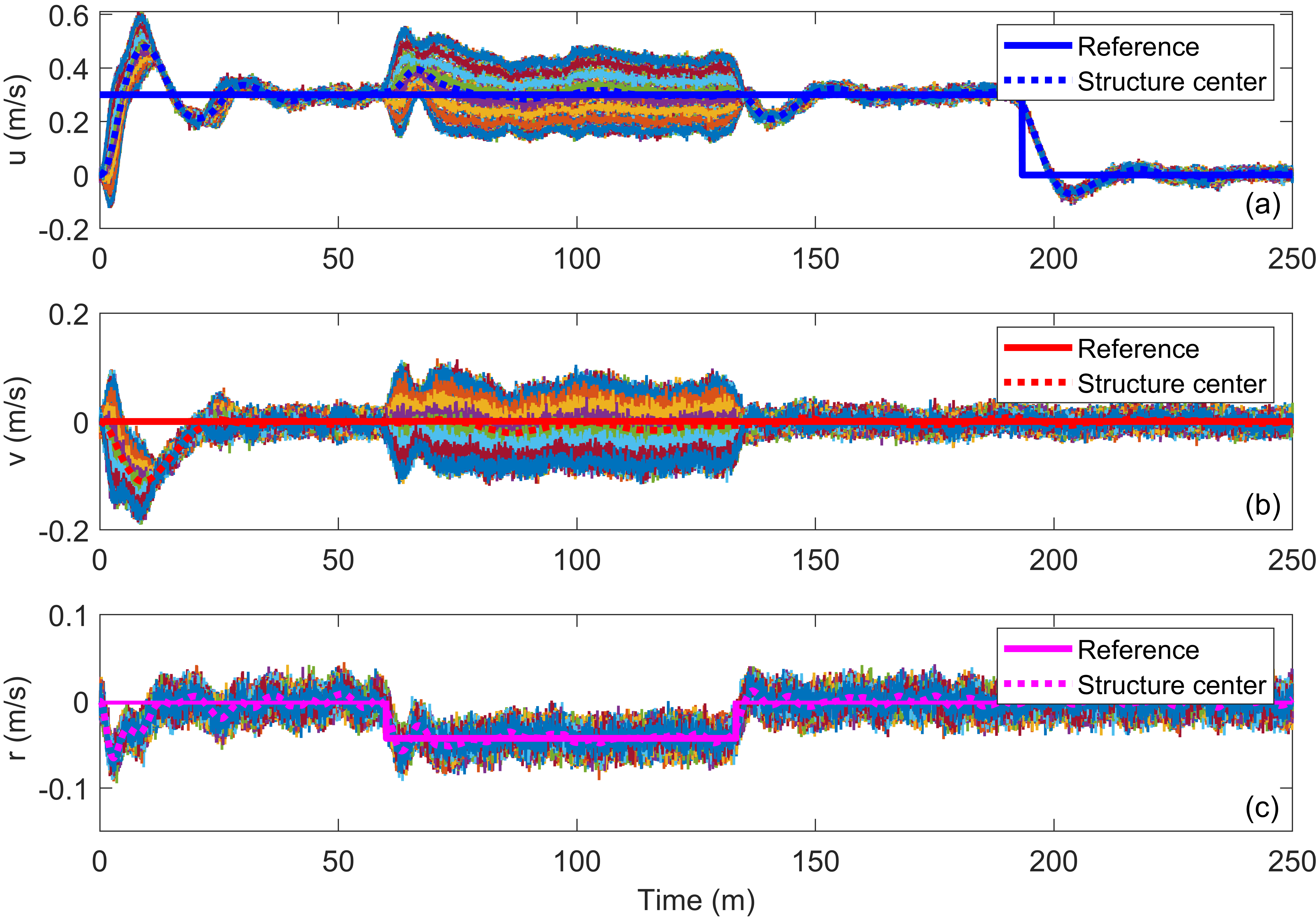}
    \caption{Measured linear (a), (b) and angular (c) velocities by different robots during the simulation. The thin colored lines are the noisy measured velocities of the followers. }
    \label{SimulationSpeed}
\end{figure}

\begin{figure}[htb]
    \centering
    \includegraphics[width=1.0\linewidth] {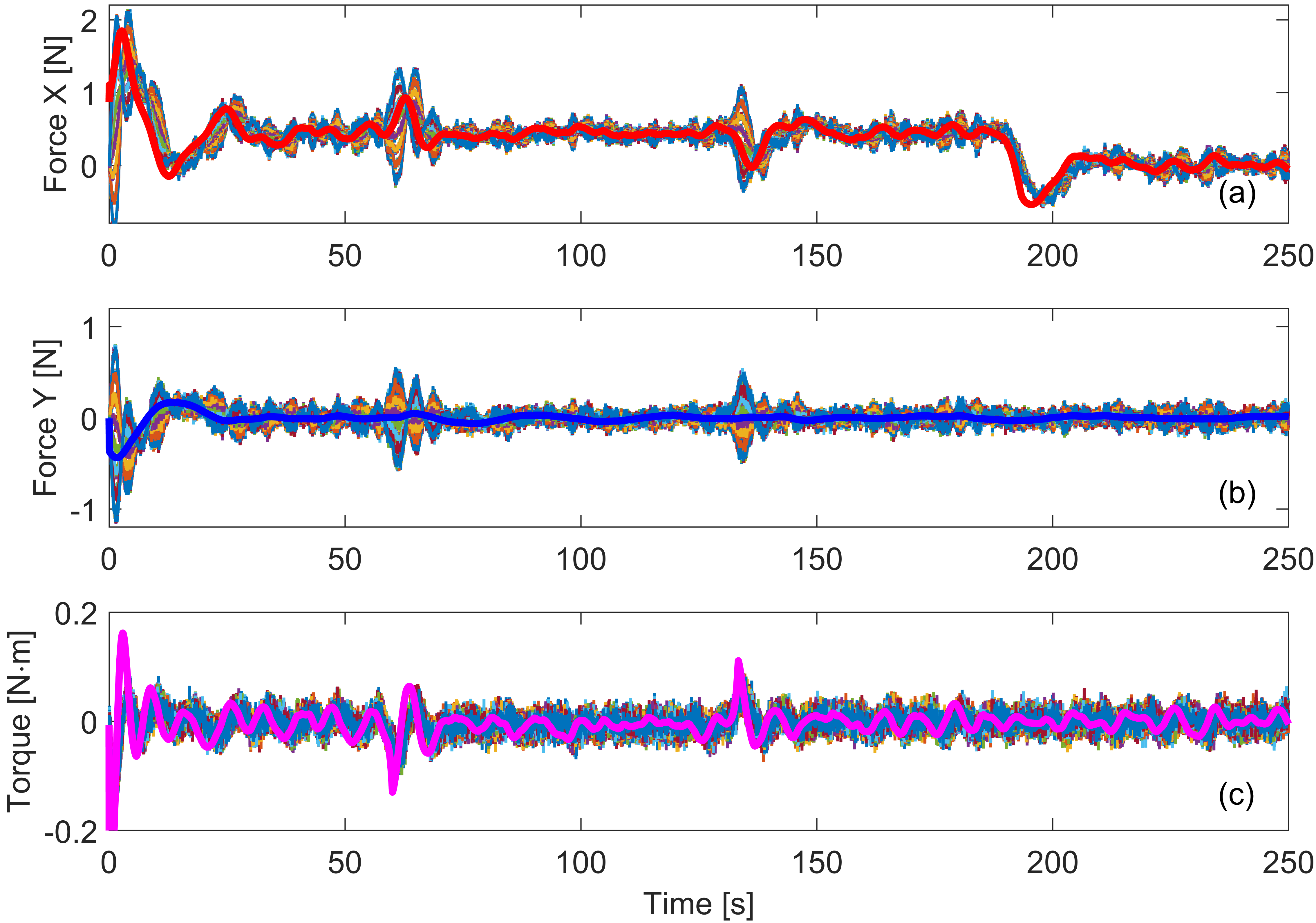}
    \caption{Forces (a), (b) and torques (c) during the simulation. The bold lines are the input force/torque of the leader robot which are generated by the NMPC. Other colored lines are the forces/torque of the followers. }
    \label{SimulationForce}
\end{figure}

Fig. \ref{SimulationTrajectoryandOrientation} and \ref{SimulationSpeed}  shows that the modular structure successfully tracks the desired trajectories, orientations, and velocities at the same time without direct communication among the robots.
It can be observed that the measured linear velocities of the robots are clearly different when the structure is rotating from 60 s to 133 s, verifying the use of local measurements. Fig. \ref{SimulationForce} shows the forces and torques of the robots during the simulation.
It is evident in Fig. \ref{SimulationForce} that both the forces and torques of the followers converge to those of the leader when the force and torque of the leader are changing,  as proved in Section \ref{DistributedFollowerController}.
Compared to the controllers in \cite{ZijianWangIJRR2016, ZijianWang7402230, ZijianWang7487163}, our controller  is able to accurately track the trajectory, orientation, linear and angular velocities of the structure simultaneously. Moreover, compared to the controller in \cite{wang2018ouijabots}, our controller  does not need the relative position of the structure and uses only the local measurements to achieve the force and torque alignment.
Furthermore, it is notable that tuning the NMPC weight parameters is straightforward because each weight parameter has an explicit physical meaning. By contrast, the PID controllers always require troublesome parameter tuning.
\section{Experiments}
\begin{figure}[htb]
    \centering
    \includegraphics[width=1.0\linewidth] {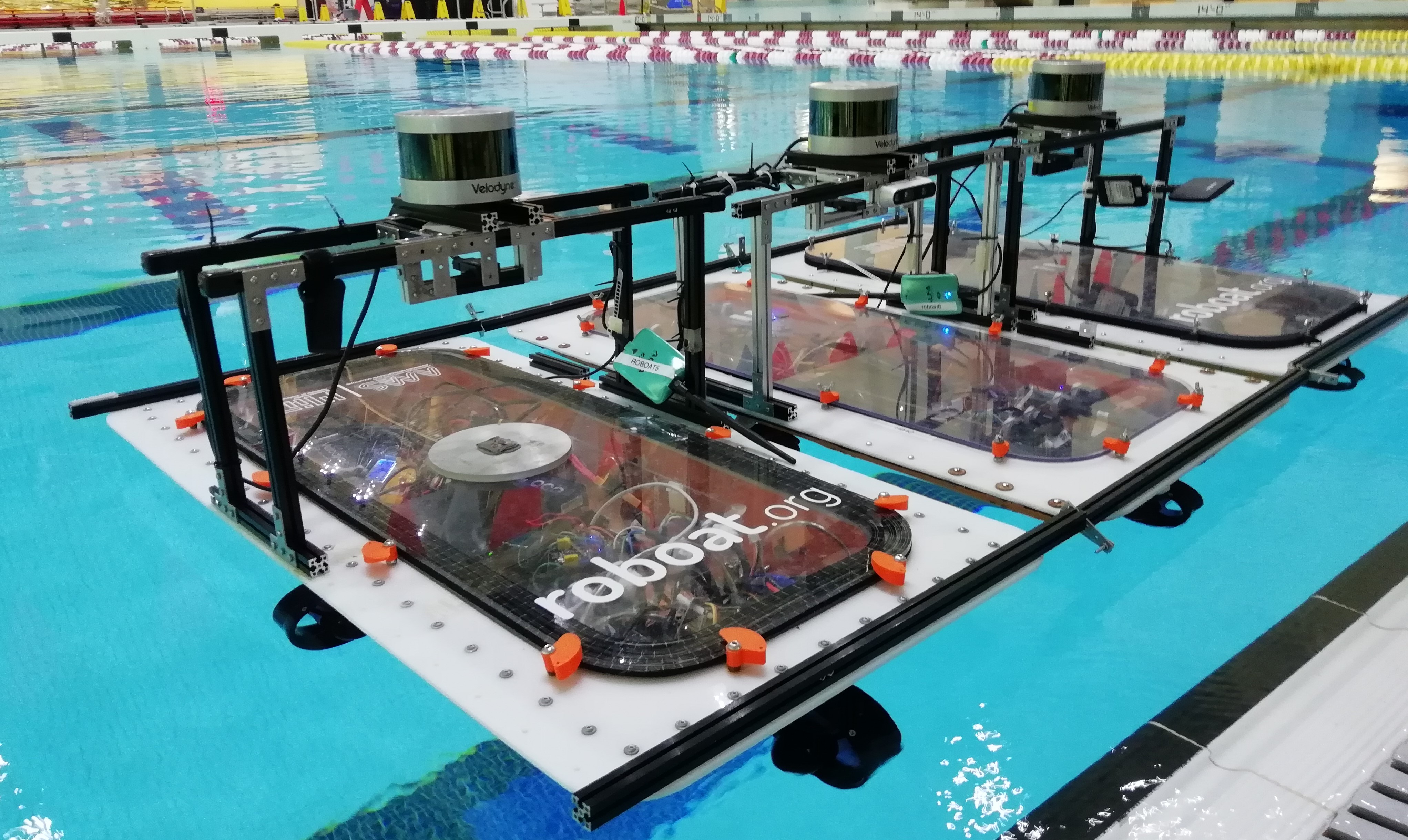}
    \caption{Picture of the floating modular structure consisting of three robot modules in parallel configuration. }
    \label{Threeroboats}
\end{figure}

In this section, the effectiveness and robustness of the proposed cooperative controllers are verified through experiments with custom-built robots.
We built three Roboats for the experiments in this study, as shown in Fig. \ref{Threeroboats}.
Each Roboat is able to estimate its states including the position, heading, velocity and acceleration by its onboard sensors. During the experiments, only the leader knows the desired trajectory, heading and velocities of the structure while each follower run a consensus-based controller (\ref{DistributedControlLawWithLocalMeasurement}) to match the forces and torque of the leader's using only onboard sensors as feedback.
In particular, two configurations are studied in the experiments. One configuration is three robots are connected in series, the other one is three robots are connected in parallel. The experiment was conducted in 12 $\times$ 6 m arena within a swimming pool. The position, heading angle, force, torque, velocity and acceleration measurements of the robots are recorded at 10 Hz during the experiments and then uploaded for analysis. All the experiments are run 5 times to ensure repeatability.

\begin{figure}[htb]
    \centering
    \includegraphics[width=1.0\linewidth] {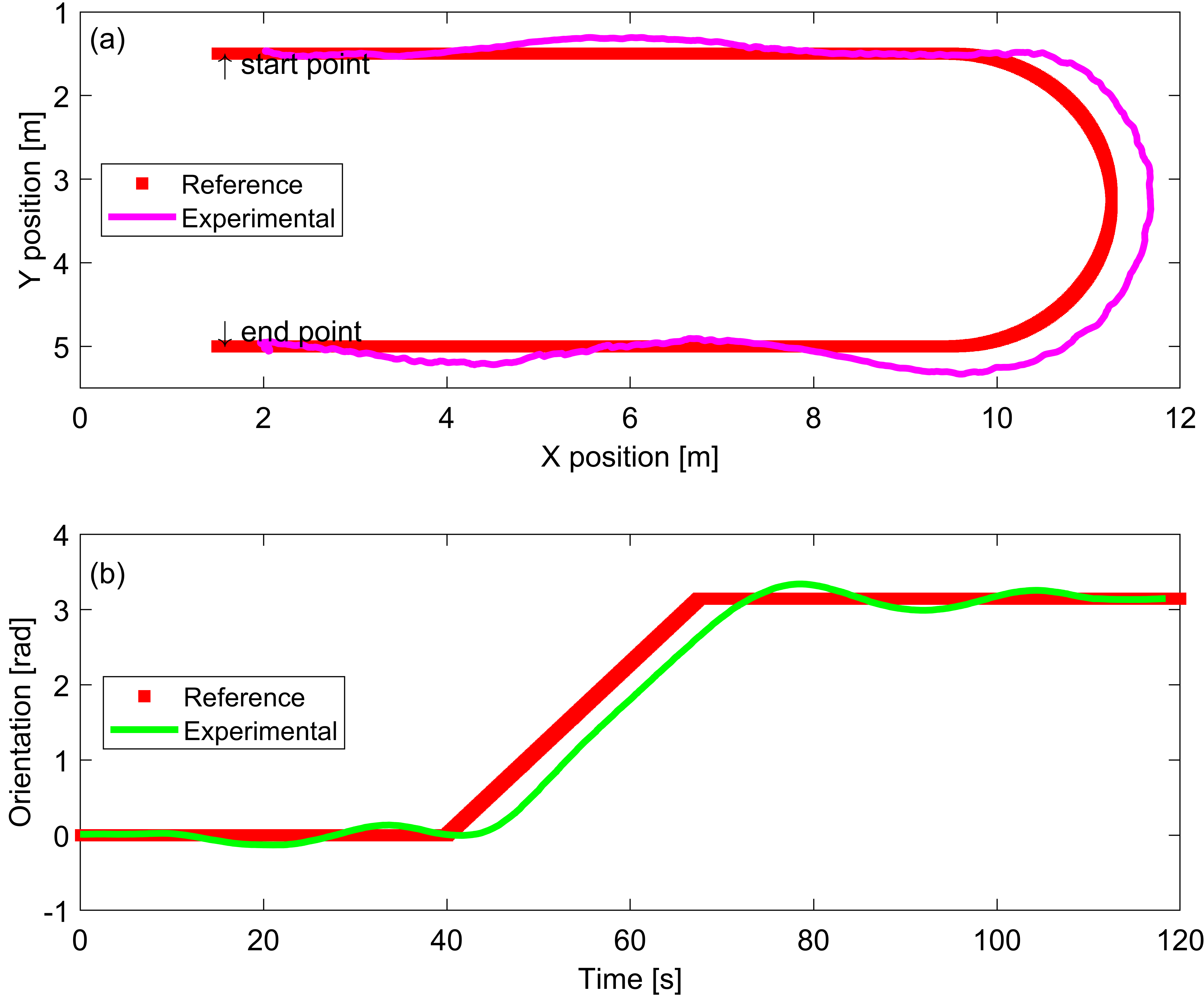}
    \caption{Overall trajectory and orientation tracking of the floating structure with three robots connected in series in the experiment. }
    \label{TrajectoryAndOrientationExperimentCascade}
\end{figure}

\begin{figure}[htb]
    \centering
    \includegraphics[width=1.0\linewidth] {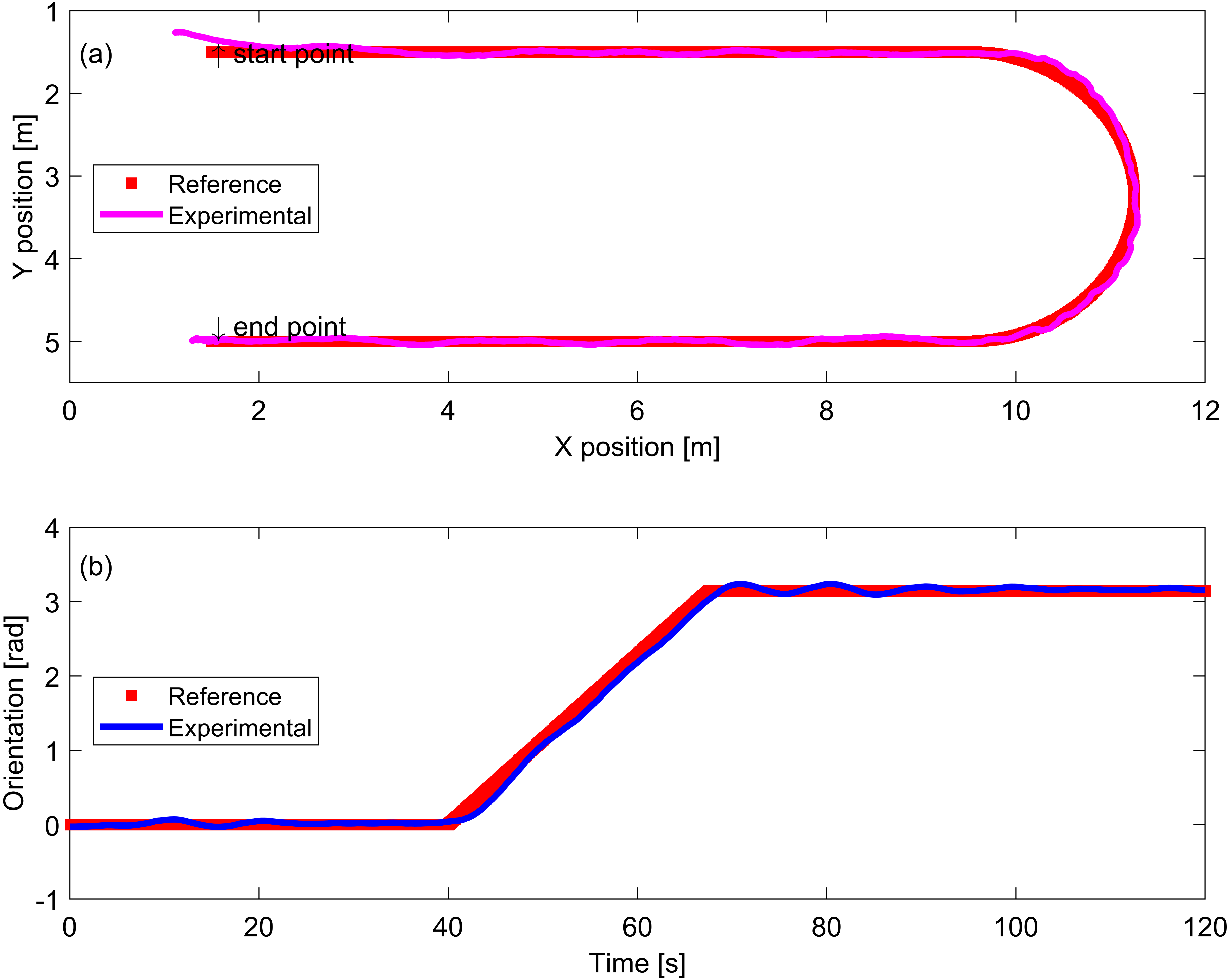}
    \caption{Overall trajectory and orientation tracking of the floating structure with three robots connected in parallel in the experiment. }
    \label{TrajectoryAndOrientationExperimentParallel}
\end{figure}

\begin{figure}[htb]
    \centering
    \includegraphics[width=1.0\linewidth] {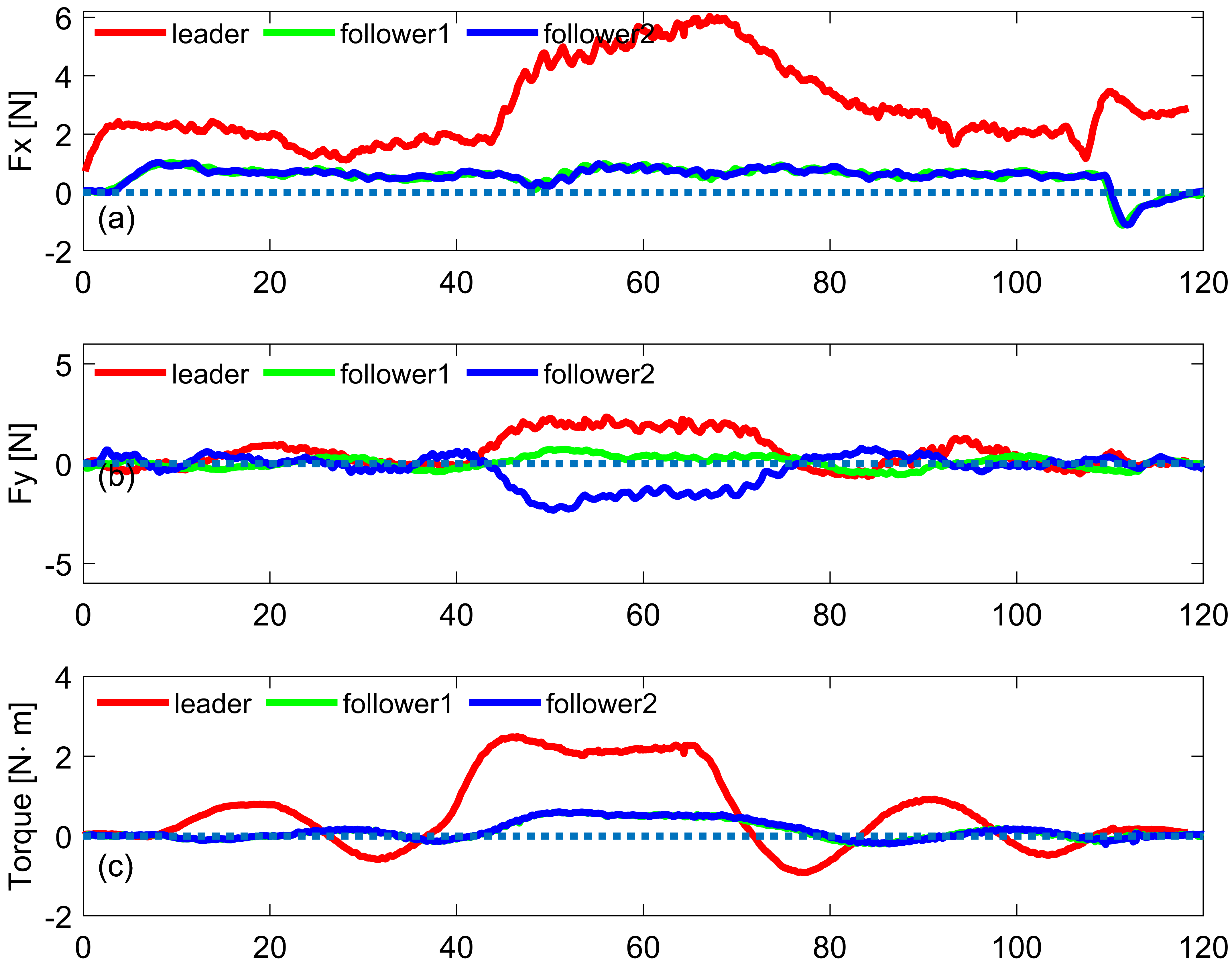}
    \caption{Forces and torques of the robots during the experiments with three robots connected in series. }
    \label{ForceTrackingExperimentCascade}
\end{figure}

\begin{figure}[htb]
    \centering
    \includegraphics[width=1.0\linewidth] {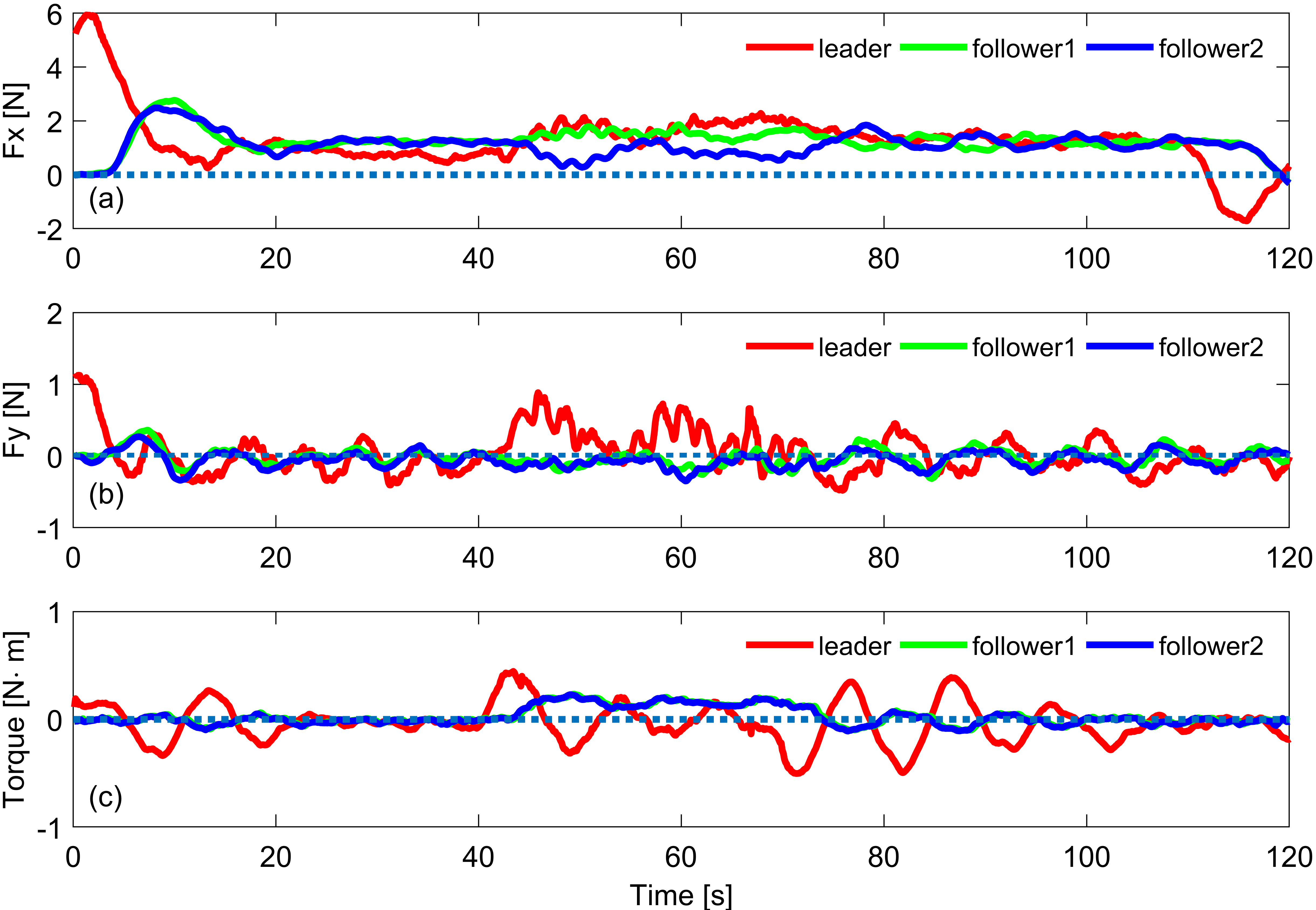}
    \caption{Forces and torques of the robots during the experiments with three robots connected in parallel. }
    \label{ForceTrackingExperimentParallel}
\end{figure}

Fig. \ref{TrajectoryAndOrientationExperimentCascade} and \ref{TrajectoryAndOrientationExperimentParallel} illustrate the tracking performance of the cooperative controllers when the floating structure is in series and in parallel, respectively.
It can be seen that the coordinated group is able to track various trajectories and orientations in both configurations. We can see that the cooperative controller performs  better when the group is configured in parallel compared with the tandem configuration. This is possibly due to the fact that the local measurements of the robots in tandem configuration have more discrepancies with respect to the center of the structure.
 Fig. \ref{ForceTrackingExperimentCascade} and \ref{ForceTrackingExperimentParallel} illustrate the forces and torques of all the three robots in series and parallel configuration, respectively.
From Fig. \ref{ForceTrackingExperimentCascade}(a) and \ref{ForceTrackingExperimentParallel}(a) we can clearly see that the magnitudes of followers' forces contribute positively, indicating that the follower robots help the leader in the process. Moreover, we can see from Fig. \ref{ForceTrackingExperimentParallel}(a) that the followers contribute almost equally with the leader, implying that the cooperation is very efficient. Furthermore, similar to the simulation in Fig. \ref{SimulationForce}, there is a slight time delay between the forces/torques of the leader and those of the followers, because it takes time for the followers to converge. In addition, from Fig. \ref{ForceTrackingExperimentCascade}(c) and \ref{ForceTrackingExperimentParallel}(c), we can also see that the torques of the followers also contribute positively especially when the floating structure is rotating from 40 s to 67 s, which indicates that the followers collaborated  with the leader to rotate the structure.

\section{Conclusion and Future Work}
In this paper, we have developed cooperative algorithms that enable multiple connected surface vessels to form a floating modular structure and cooperatively track desired trajectories, orientations and velocities on the water without any cross-robot communication.   An NMPC strategy is first formulated for the leader which is subject to control inputs and dynamics constraints. A dynamical model for the floating modular structure with $N$ modules is then derived. This model-based leader controller is able to adapt to different sizes and configurations of the group without parameter re-tuning.  We further propose a decentralized force and torque controller for the followers, which requires only local measurements (e.g., velocities and accelerations).
Furthermore, we build three autonomous modular vessels, which are capable of holonomic motions and onboard state estimation using an extended Kalman filter.
We have theoretically proved the convergence of the cooperative controller.  Extensively simulations and experiments with the robots have verified that the connected modules can  successfully coordinate their force and torque to complete the tracking tasks relying on their local measurements, rather than using explicit communication.

In the future, we can use machine learning to online estimate the key parameters the robots which are now roughly calculated  in Section \ref{DynamicsOfFloatingModularStructure}. We are also interested in adaptive controllers that allow for the dynamics change to the structure when meaningful objects are placed on the structure, such as the work in \cite{culbertson2018decentralized}. Moreover, we are extending our approach to realistic outdoor water environments where large disturbances such as currents and waves exist.
\bibliographystyle{IEEEtran}

\bibliography{CollectiveTransportIROS2020FINAL}

\begin{thebibliography}{10}
\providecommand{\url}[1]{#1}
\csname url@samestyle\endcsname
\providecommand{\newblock}{\relax}
\providecommand{\bibinfo}[2]{#2}
\providecommand{\BIBentrySTDinterwordspacing}{\spaceskip=0pt\relax}
\providecommand{\BIBentryALTinterwordstretchfactor}{4}
\providecommand{\BIBentryALTinterwordspacing}{\spaceskip=\fontdimen2\font plus
\BIBentryALTinterwordstretchfactor\fontdimen3\font minus
  \fontdimen4\font\relax}
\providecommand{\BIBforeignlanguage}[2]{{%
\expandafter\ifx\csname l@#1\endcsname\relax
\typeout{** WARNING: IEEEtran.bst: No hyphenation pattern has been}%
\typeout{** loaded for the language `#1'. Using the pattern for}%
\typeout{** the default language instead.}%
\else
\language=\csname l@#1\endcsname
\fi
#2}}
\providecommand{\BIBdecl}{\relax}
\BIBdecl

\bibitem{couzin2003self}
I.~D. Couzin and J.~Krause, ``Self-organization and collective behavior in
  vertebrates,'' \emph{Advances in the Study of Behavior}, vol.~32, pp. 1--75,
  2003.

\bibitem{Mlot7669}
N.~J. Mlot, C.~A. Tovey, and D.~L. Hu, ``Fire ants self-assemble into
  waterproof rafts to survive floods,'' \emph{Proceedings of the National
  Academy of Sciences}, vol. 108, no.~19, pp. 7669--7673, 2011.

\bibitem{Alonso-Mora2018}
J.~Alonso-Mora, E.~Montijano, T.~N{\"a}geli, O.~Hilliges, M.~Schwager, and
  D.~Rus, ``Distributed multi-robot formation control in dynamic
  environments,'' \emph{Autonomous Robots}, 2018.

\bibitem{Olfati-Saber-733}
R.~Olfati-Saber and R.~Murray, ``Consensus problems in networks of agents with
  switching topology and time-delays,'' \emph{IEEE Transactions on Automatic
  Control}, vol.~49, no.~9, pp. 1520 -- 1533, 2004.

\bibitem{Rus5569030}
K.~{Gilpin} and D.~{Rus}, ``Modular robot systems,'' \emph{IEEE Robotics
  Automation Magazine}, vol.~17, no.~3, pp. 38--55, Sep. 2010.

\bibitem{Murata4141035}
S.~{Murata} and H.~{Kurokawa}, ``Self-reconfigurable robots,'' \emph{IEEE
  Robotics Automation Magazine}, vol.~14, no.~1, pp. 71--78, March 2007.

\bibitem{Yim4141032}
M.~{Yim}, W.~{Shen}, B.~{Salemi}, D.~{Rus}, M.~{Moll}, H.~{Lipson},
  E.~{Klavins}, and G.~S. {Chirikjian}, ``Modular self-reconfigurable robot
  systems [grand challenges of robotics],'' \emph{IEEE Robotics Automation
  Magazine}, vol.~14, no.~1, pp. 43--52, March 2007.

\bibitem{Daudelineaat4983}
\BIBentryALTinterwordspacing
J.~Daudelin, G.~Jing, T.~Tosun, M.~Yim, H.~Kress-Gazit, and M.~Campbell, ``An
  integrated system for perception-driven autonomy with modular robots,''
  \emph{Science Robotics}, vol.~3, no.~23, 2018. [Online]. Available:
  \url{https://robotics.sciencemag.org/content/3/23/eaat4983}
\BIBentrySTDinterwordspacing

\bibitem{Raymond0278364913501212}
R.~Oung and R.~D’Andrea, ``The distributed flight array: Design,
  implementation, and analysis of a modular vertical take-off and landing
  vehicle,'' \emph{The International Journal of Robotics Research}, vol.~33,
  no.~3, pp. 375--400, 2014.

\bibitem{Kumar8461014}
D.~{Saldaña}, B.~{Gabrich}, G.~{Li}, M.~{Yim}, and V.~{Kumar}, ``Modquad: The
  flying modular structure that self-assembles in midair,'' in \emph{2018 IEEE
  International Conference on Robotics and Automation (ICRA)}, May 2018, pp.
  691--698.

\bibitem{Baldassarre4067066}
G.~{Baldassarre}, V.~{Trianni}, M.~{Bonani}, F.~{Mondada}, M.~{Dorigo}, and
  S.~{Nolfi}, ``Self-organized coordinated motion in groups of physically
  connected robots,'' \emph{IEEE Transactions on Systems, Man, and Cybernetics,
  Part B (Cybernetics)}, vol.~37, no.~1, pp. 224--239, Feb 2007.

\bibitem{paulos2015automated}
J.~Paulos, N.~Eckenstein, T.~Tosun, J.~Seo, J.~Davey, J.~Greco, V.~Kumar, and
  M.~Yim, ``Automated self-assembly of large maritime structures by a team of
  robotic boats,'' \emph{IEEE Transactions on Automation Science and
  Engineering}, vol.~12, no.~3, pp. 958--968, 2015.

\bibitem{Tuci2018a}
E.~Tuci, M.~H.~M. Alkilabi, and O.~Akanyeti, ``Cooperative object transport in
  multi-robot systems: A review of the state-of-the-art,'' \emph{Frontiers in
  Robotics and AI}, vol.~5, p.~59, 2018.

\bibitem{Rus525802}
D.~{Rus}, B.~{Donald}, and J.~{Jennings}, ``Moving furniture with teams of
  autonomous robots,'' in \emph{Proceedings 1995 IEEE/RSJ International
  Conference on Intelligent Robots and Systems. Human Robot Interaction and
  Cooperative Robots}, vol.~1, Aug 1995, pp. 235--242 vol.1.

\bibitem{Mellinger2013}
D.~Mellinger, M.~Shomin, N.~Michael, and V.~Kumar, \emph{Cooperative Grasping
  and Transport Using Multiple Quadrotors}.\hskip 1em plus 0.5em minus
  0.4em\relax Berlin, Heidelberg: Springer Berlin Heidelberg, 2013, pp.
  545--558.

\bibitem{ZijianWang7402230}
Z.~{Wang} and M.~{Schwager}, ``Multi-robot manipulation with no communication
  using only local measurements,'' in \emph{2015 54th IEEE Conference on
  Decision and Control (CDC)}, Dec 2015, pp. 380--385.

\bibitem{ZijianWangIJRR2016}
Z.~Wang and M.~Schwager, ``Force-amplifying {N-robot} transport system
  ({Force-ANTS}) for cooperative planar manipulation without communication,''
  \emph{The International Journal of Robotics Research}, vol.~35, no.~13, pp.
  1564--1586, 2016.

\bibitem{Franchi8476991}
A.~{Franchi}, A.~{Petitti}, and A.~{Rizzo}, ``Distributed estimation of state
  and parameters in multi-agent cooperative manipulation,'' \emph{IEEE
  Transactions on Control of Network Systems}, pp. 1--1, 2018.

\bibitem{Michael7989533}
M.~{Corah} and N.~{Michael}, ``Active estimation of mass properties for safe
  cooperative lifting,'' in \emph{2017 IEEE International Conference on
  Robotics and Automation (ICRA)}, May 2017, pp. 4582--4587.

\bibitem{wang2016multi}
Z.~Wang and M.~Schwager, ``Multi-robot manipulation without communication,'' in
  \emph{Distributed autonomous robotic systems}.\hskip 1em plus 0.5em minus
  0.4em\relax Springer, 2016, pp. 135--149.

\bibitem{ZijianWang7487163}
Z.~{Wang} and M.~{Schwager}, ``Kinematic multi-robot manipulation with no
  communication using force feedback,'' in \emph{2016 IEEE International
  Conference on Robotics and Automation (ICRA)}, May 2016, pp. 427--432.

\bibitem{Esposito4543414}
J.~{Esposito}, M.~{Feemster}, and E.~{Smith}, ``Cooperative manipulation on the
  water using a swarm of autonomous tugboats,'' in \emph{2008 IEEE
  International Conference on Robotics and Automation}, May 2008, pp.
  1501--1506.

\bibitem{Hajieghrary8430951}
H.~{Hajieghrary}, D.~{Kularatne}, and M.~A. {Hsieh}, ``Differential geometric
  approach to trajectory planning: Cooperative transport by a team of
  autonomous marine vehicles,'' in \emph{2018 Annual American Control
  Conference (ACC)}, June 2018, pp. 858--863.

\bibitem{Hajieghrary8206033}
------, ``Cooperative transport of a buoyant load: A differential geometric
  approach,'' in \emph{2017 IEEE/RSJ International Conference on Intelligent
  Robots and Systems (IROS)}, Sep. 2017, pp. 2158--2163.

\bibitem{wang2018ouijabots}
Z.~Wang, G.~Yang, X.~Su, and M.~Schwager, ``Ouijabots: Omnidirectional robots
  for cooperative object transport with rotation control using no
  communication,'' in \emph{Distributed Autonomous Robotic Systems}.\hskip 1em
  plus 0.5em minus 0.4em\relax Springer, 2018, pp. 117--131.

\bibitem{wang2019cooperative}
W.~Wang, L.~Mateos, Z.~Wang, K.~W. Huang, M.~Schwager, C.~Ratti, and D.~Rus,
  ``Cooperative control of an autonomous floating modular structure without
  communication,'' in \emph{2019 International Symposium on Multi-Robot and
  Multi-Agent Systems (MRS)}.\hskip 1em plus 0.5em minus 0.4em\relax IEEE,
  2019, pp. 44--46.

\bibitem{WeiICRA2018}
W.~Wang, L.~Mateos, S.~Park, P.~Leoni, B.~Gheneti, F.~Duarte, C.~Ratti, and
  D.~Rus, ``Design, modeling, and nonlinear model predictive tracking control
  of a novel autonomous surface vehicle,'' in \emph{Proc. 2018 IEEE Int. Conf.
  Robot. Autom}, 2018, pp. 6189--6196.

\bibitem{bishop2001introduction}
G.~Bishop, G.~Welch \emph{et~al.}, ``An introduction to the kalman filter,''
  \emph{Proc of SIGGRAPH, Course}, vol.~8, no. 27599-23175, p.~41, 2001.

\bibitem{BiberStrasser2006}
P.~Biber and W.~Strasser, ``The normal distributions transform: a new approach
  to laser scan matching,'' in \emph{IEEE/RSJ International Conference on
  Intelligent Robots and Systems (IROS)}, 2003, pp. 2743 -- 2748.

\bibitem{moore2016generalized}
T.~Moore and D.~Stouch, ``A generalized extended kalman filter implementation
  for the robot operating system,'' in \emph{Intelligent Autonomous Systems
  13}.\hskip 1em plus 0.5em minus 0.4em\relax Springer, 2016, pp. 335--348.

\bibitem{WeiIROS2019}
W.~{Wang}, B.~{Gheneti}, L.~A. {Mateos}, F.~{Duarte}, C.~{Ratti}, and D.~{Rus},
  ``Roboat: An autonomous surface vehicle for urban waterways,'' in \emph{2019
  IEEE/RSJ International Conference on Intelligent Robots and Systems (IROS)},
  Nov 2019, pp. 6340--6347.

\bibitem{Fossen285}
T.~I. Fossen, \emph{Guidance and control of ocean vehicles}.\hskip 1em plus
  0.5em minus 0.4em\relax West Sussex PO19 1UD, England: John Wiley \& Sons
  Ltd, 1994.

\bibitem{nakayama2018introduction}
Y.~Nakayama, \emph{Introduction to fluid mechanics}.\hskip 1em plus 0.5em minus
  0.4em\relax Butterworth-Heinemann, 2018.

\bibitem{houska2012acado}
B.~Houska, H.~Ferreau, F.~Logist, and M.~Diehl, ``Acado toolkit-an automatic
  control and dynamic optimization,'' 2012, optimization in Engineering Center
  (OPTEC), http://acado.github.io/.

\bibitem{culbertson2018decentralized}
P.~Culbertson and M.~Schwager, ``Decentralized adaptive control for
  collaborative manipulation,'' in \emph{2018 IEEE International Conference on
  Robotics and Automation (ICRA)}, 2018, pp. 278--285.

\end{thebibliography}
\end{document}